\documentclass[10pt, letterpaper, conference]{IEEEtran}
\IEEEoverridecommandlockouts
\usepackage{cite}
\usepackage{amsmath,amssymb,amsfonts}
\usepackage{algorithmic}
\usepackage[ruled,vlined]{algorithm2e}
\usepackage{graphicx}
\usepackage{textcomp}
\usepackage{xcolor}
\usepackage{pgfplots}
\usepackage{pgfplotstable}
\pgfplotsset{compat=1.7}
\usepackage{tikz}
\usepackage{graphicx}
\usepackage{caption}
\usepackage{subcaption}
\usepackage[margin=1in]{geometry}
\def\BibTeX{{\rm B\kern-.05em{\sc i\kern-.025em b}\kern-.08em
    T\kern-.1667em\lower.7ex\hbox{E}\kern-.125emX}}
\usepackage[hyphens]{url}
\usepackage{hyperref}

\graphicspath{{figures/}}

\begin{document}

\title{\LARGE \textbf{Online Indoor Localization Using DOA of Wireless Signals}}

\author{Ehsan Latif \and Ramviyas Parasuraman
\thanks{The authors are from the Heterogeneous Robotics Lab, Department of Computer Science, University of Georgia, Athens, GA 30602, USA. 

Email: {\it \{ehsan.latif,ramviyas\}@uga.edu}. }
}

\maketitle

\begin{abstract}
Localization of a wireless mobile device or a robot in indoor and GPS-denied environments is a difficult problem, particularly in dynamic scenarios where traditional cameras and LIDAR-based alternative sensing and localization modalities may fail.
We propose a method for estimating the location of a mobile robot in relation to static wireless sensor nodes (WSN) deployed in the environment. The method employs a novel particle filter that updates its weights using a Gauss probability over Direction of Arrival (DOA) estimate in conjunction with the mobile robot's mobility model.
We evaluate and validate the proposed method in terms of accuracy and computational efficiency through extensive simulations and public real-world measurement datasets, comparing with standard state-of-the-art localization approaches. The results show considerably high meter-level localization accuracy balanced by the high computational efficiency, enabling it to use online without a need for a dedicated offline phase as in typical fingerprint-based localization algorithms. 
\end{abstract}


\section{Introduction}
\label{sec:intro}

Location information is critical for mobile robot and wireless device operations, and device localization has been a significant challenge task, especially in indoor environments. For most outdoor location-based services, a Global Positioning System (GPS) could provide adequate positioning accuracy. However, because the GPS satellite signal cannot penetrate most buildings, it cannot provide sufficient position accuracy indoors \cite{GPSindoor}.
Furthermore, indoor, GPS-denied, or dynamically changing environments pose additional challenges for mobile robot (vehicle) localization \cite{survey}.
Indoor localization has emerged as one of the most critical components of robotics, automation, and wireless systems. One fundamental requirement is to provide an accurate and efficient localization system in a real-time (online) manner \cite{RFID2021survey7}.

Further, indoor localization is generally more difficult to achieve than outdoor positioning (via GPS) because the implementation of technologies used indoors may necessitate some additional infrastructure. Some example scenarios of indoor localization are searching for specific wireless devices, locating rooms within an institution's building, orienting oneself within the hospital complex, locating specific stores within the mall, navigating easily to the airport, museum, and so on \cite{survey8}.

For consumer electronics, wireless technologies such as Wi-Fi and Bluetooth are the most extensively utilized for indoor WLANs. The ubiquitous availability of RSSI measurement on such inexpensive commercial devices is the Received Signal Strength Indicator (RSSI) measured from an Access Point (AP) or a Wireless Sensor Node (WSN). This RSSI values can be used in various applications, including localization \cite{yang2020trilateration,jianyong2014rssi,parashar2020particle}, control \cite{luo2019multi}, and communication optimization \cite{parasuraman2013spatial,parasuraman2018kalman}.

In indoor positioning and indoor navigation systems research, most of the existing work focuses on creating a fingerprinting database and using the fingerprinted data along with current real-time RSSI data to position the wireless device using a supervised machine learning algorithm \cite{fingrfloc,sadowski2020memoryless}. However, these fingerprinting-based approaches require a dedicated offline phase, in addition to the limitation of generalization where they can be employed only for the specific environment where fingerprinted \cite{tao2018novel}.
Therefore, there is a need for the real-time (online) capability of the localization method without a dedicated fingerprinting process \cite{survey}.

In recent years, alternative sensing technologies such as LIDAR, Infrared, inertial measurement units (IMU), camera image-based visual localization, and their fusion with wireless signal-based positioning systems have been exploited for obtaining accurate localization of mobile devices \cite{jianyong2014rssi,canedo2016particle}. However, these technologies are expensive or suffer from various limitations of the alternate sensor modalities, such as the requirement of proper lighting conditions for the camera, structured surfaces for LIDAR, gross accumulation errors of IMU.
To enable better fusion with alternate sensing modalities, RSSI-based positioning algorithms should provide high accuracy as well as high computational efficiency (online) without the need for a dedicated training phase or tweaking to a specific environment.

\begin{figure}[t]
\centering
 \includegraphics[width=\columnwidth]{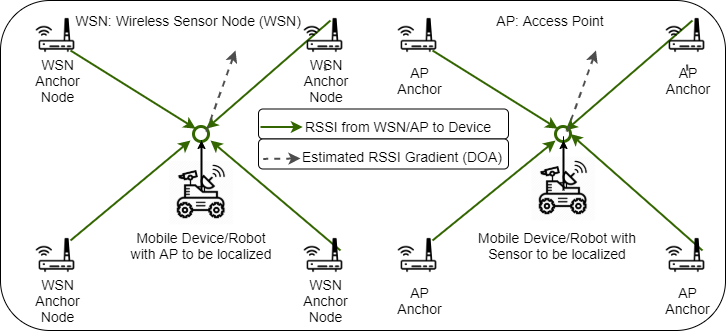}
 \caption{Overview of the proposed device localization approach with WSNs (left) and APs (right) as anchors.}
 \label{fig:overview}
 \vspace{-4mm}
\end{figure}

To address these gaps, we propose a novel online indoor localization method that uses the Direction of Arrival (DOA) estimated through RSSI of wireless signals (instead of directly using RSSI values). This metric is then combined with the well-known and robust Bayesian framework of particle filters (PF) to perform mobile robot localization. We design two different equally-capable schemes to demonstrate our RSSI-based DOA estimation for localization: 1) using RSSI measurements from three or more geometrically-aligned Wireless Access Points (APs) deployed in the infrastructure; 2) using Wireless Sensors Nodes (WSN) spatially distributed in the environment in a specific fashion communicating their RSSI values to the wireless Access Point (AP) sitting on the mobile robot (device). See Fig.~\ref{fig:overview} for an overview of the proposed method on the WSN-based and the AP-based RSSI measurement schemes.
The main contributions of this paper are outlined below.
\begin{enumerate}
   \item We propose a novel model-based PF-DOA indoor localization algorithm by using a particle filter updated based on the DOA information extracted from RSS data in two different schemes.
   \item We verify the proposed method through extensive numerical simulations and real-world datasets available from the literature and analyze the methods from the perspective of impacts by different regions, wireless technologies, and channels.
    \item We validate the accuracy and efficiency of the proposed method compared to recent non-fingerprinting methods from the literature such as the trilateration \cite{yang2020trilateration,fundamentalTrilateration1996}, weighted centroid \cite{weightedCentroid2014survey}, differential RSS \cite{podevijn2018comparison}, and Markov grid (or expectation maximization \cite{measRSS}) approaches. 
    \item We open source the implementation of the algorithm in Python, including the compared methods, via GitHub\footnote{\url{https://github.com/herolab-uga/pf-doa-localization}}. The resource also includes the python scripts for all implemented methods and the simulated dataset, as well as real-world RSSI datasets for indoor localization to enable reproducibility and practicality in real-world applications.
\end{enumerate}

In our previous work \cite{parashar2020particle}, we demonstrated an online localization of a stationary AP from a mobile robot using PF-based estimation aided by DOA of wireless signals. However, in this work, we focus on mobile devices to enable broader application on different infrastructure possibilities in indoor scenarios, and we improve the PF algorithm in terms of performance and efficiency.
Specifically, the proposed method enables the fusion of the robot's motion model (or inertial odometry) with that of the Gaussian probability of DOA estimates in the PF's resampling process. Through these novelties, our method achieves superior localization accuracy while also enabling real-time efficiency compared to several state-of-the-art solutions.

\section{Related Work}
\label{sec:relatedwork}

According to the literature \cite{localization2013survey,liu2020survery4}, there are two categories of wireless localization solutions: \textit{Model-based and Survey-based}. Model-based approaches include trilateration using RSSI, triangulation using DOA, and Weighted Centroid using distance. Recent works from the literature include variants thereof, such as the filtered trilateration \cite{yang2020trilateration}, differential RSSI-based least squares estimation \cite{podevijn2018comparison}, and Expectation Maximization \cite{pajovic2015unsupervised}.
On the other hand, survey-based approaches provide high accuracy on the basis of precise fingerprints, but with high computation cost of prediction algorithms like the K-Nearest Neighbors \cite{subedi2020survey5,zhuang2016sensors}. Therefore, we focus on the model-based solution in our work.

In model-based approaches, multilateration and triangulation are the fundamental methods to predict the position of a wireless device using RSSI captured from multiple anchors/APs \cite{passafiume2016triangulation}. However, these methods suffer from co-linearity, ambiguous positioning, non-intersecting circles, etc. 
A survey \cite{zafari2019survey1} shows that in model-based techniques, the balance of accuracy and computing complexity is absent. Also, different variants of trilateration/triangulation can have different localization accuracy, resulting in inconsistency in its application. The weighted centroid method is less accurate for Non-Line-of-Sight conditions, limiting its applicability.

DOA (or Angle of Arrival) based methods achieve high localization accuracy. For instance, in \cite{arbula2020sensors}, the authors used WSN equipped with Infrared arrays to estimate the DOA of a mobile robot, which was then used to achieve indoor localization with meter-level accuracy. 
There have been few attempts in the past using bearing information along with range measurement for offline device localization, but most of the methods are survey-based and are computationally expensive.

We depart from the previous works by estimating Gaussian probability based on the difference between the DOA calculated by a geometric model and the real observations from WSNs. Imposing a geometric model of the AP/WSN distribution in the environment, we propose a DOA estimation scheme from RSSI measurements. To achieve a balance between accuracy requirements, quick online operation, and fingerprinting independence, the approach adopts a particle filter, which can be applied independent of the robot's motion model or combined with the robot's odometry, if available.
Our proposed approach is advantageous in the sense that it reduces computational complexity without embedding external hardware and uses bearing-only information (aided by RSSI measurements), and achieves high accuracy even in the presence of signal noise.

\section{Proposed Approach}
\label{sec:approach}
We look at the problem of a robot (or a wireless device) localization (self-localization) against its surroundings.
We propose two different schemes to support our method.
In the first scheme, there exist several WSNs distributed in the environment, and that the mobile robot is mounted with an AP, which can be sensed by nearby WSNs. The WSNs measure the RSSI values coming from the AP and communicate this information to the robot (assuming the measurements and communicated data are reasonably time-synchronized). 
In the second scheme, there exist multiple APs in the surroundings, which can be sensed by the wireless device on a mobile robot. The robot measures RSSI from those APs simultaneously (e.g., using \textit{iwscan} tool for Wi-Fi RSSI scan). This is similar to the Wi-Fi infrastructure in most non-residential buildings.
In the rest of this section, we use the first scheme to describe the proposed methods. Our simulations use this first scheme, while our validation on real-world datasets uses the second scheme.

\textbf{Problem Statement} Suppose that we have a wireless AP mounted on a mobile robot (whose location to be estimated) and that four fixed WSNs \(N=\{ N_1,N_2,...N_4 \}\) are placed at the corners of the bounded region (see Fig.~\ref{fig:overview} for location reference). The WSNs measures the radio signal strength (RSS) coming from the robot's AP. Using these signal values, we can estimate the DOA of the robot's vector in relation to the WSN's center. The robot \(R\) records its path along with the DOA measurements as the tuple: \( m_l = \{ x_l , y_l , DOA_l\}\), where \((x_l , y_l)\) is the location of the robot at location \(l\). The problem is to find the best estimate of the robot's location \((x_{R}^* , y_{R}^* ) \in \mathbb{R}^2\) which maximizes the probability of observing the measurement tuples when the robot is at the estimated location \(P(x_{R} , y_{R} \mid m_l , m_{l-1}, . . . , m_{l-M})\), where and \(M\) is the number of previous samples considered along the completed path trajectory so far, given that we employ an arbitrary method to estimate the DOA.

Our solution uses a particle filter-based localization of a mobile AP. Similar to \cite{li2015modified} for acoustic signals, we employ a Gaussian probability model on the wireless signal DOA estimates to calculate the weights of each random particle in the PF. This probabilistic model will weigh the quality of signals sensed by each node from \(N\) and ultimately produce an accurate robot location estimate through the PF. 
RSS can be modeled as a vector with two components, and the gradient with respect to the center of the robot can be represented as \(\vec{g} =[g_x,g_y]\). Using the central finite difference method \cite{parasuraman2013spatial}, the gradient can be calculated as follows:
\begin{equation}
    g_x=\frac{S_{N3} - S_{N2}}{2\Delta_{X}} + \frac{S_{N4} - S_{N1}}{2\Delta_{X}},
\end{equation}
\begin{equation}
    g_y=\frac{S_{N3} - S_{N4}}{2\Delta_{Y}} + \frac{S_{N2} - S_{N1}}{2\Delta_{Y}}
\end{equation}

Here, \(\Delta_{X}\) is the distance between the wireless sensor's antennas along the x-axis, \(\Delta_{Y}\) is the distance between the wireless sensor's antennas along the y-axis, and \(S_{N1}, S_{N2}, S_{N3}\), and \(S_{N4}\) are the RSS values at nodes \(N=\{ N_1,N_2,...N_4 \}\), respectively on the mobile robot, measured at the current path location. 
\begin{equation}
    DOA_l=\arctan(\frac{g_y}{g_x}) .
\end{equation}
This provides the DOA of the wireless signal at a position along the path \(l\) using the RSS gradient.
We can then suppress the noise of the calculated DOA by using the exponential weighted moving normal. In which we are considering a fixed window \(k\) of previous positions of the robot along its path. In this way, the sampled DOA at a path area \(l\) with \(K - 1\) past positions included in the window can be composed as:
\begin{equation}
    \overline{x}_k =\frac{1}{0.99^0+0.99^1+...+0.99^{k-1}} ,
\end{equation}
\begin{equation}
    \widetilde{DOA_l} = \overline{x}_k \times \sum_{n=0}^{k-1} 0.99^n \times DOA_{l-n} .
\end{equation}

The error between actual RSS (AS) for all wireless sensors at a potential candidate position \(l\) of mobile robot with coordinates \((x_c, y_c)\) to the perceived RSS values for each sensor, can be calculated as follows:
\begin{equation}
    err_l^{ws}=\sum_{j=1}^{N}(AS_l^j-S_l^j) .
    \label{equation3}
\end{equation}
Now, we use the Gaussian probability formula (similar to \cite{parashar2020particle,li2015modified}) on this error to calculate the probability of the \(i_{th}\) candidate location of the particle \( q_i = (x_i, y_i, w_i), i \in (1, ..N)\), where \(N\) is the number of particles in the PF and \(w_i\) is the weight of each particle calculated over a set of previous path samples as:
\begin{equation}
    w_i  \propto P_l(q_i) = \prod_{k=0}^{M-1} \bigg[\frac{1}{\sigma\sqrt{2\pi}}e^{-\frac{(err_l^{t-k})^2}{2\sigma^2}}\bigg].
    \label{eqn:weights}
\end{equation}

There is an intrinsic angular inaccuracy in each DOA degree that is analyzed. \(\sigma\) represents the fluctuation (deviation) of this error, which is anticipated to be known because we know the correctness of the technique being used to assess the DOA. We use the product of the Gauss likelihood of DOA error over \(M - 1\) prior robot positions (imitating geographically scattered samples) so that the sifted DOA from earlier path locations can be used in the same way as readings from many sensors. 

This component of the DOA probability is used to calculate the weights of the particles in the PF, which is then employed in the resampling procedure in the next PF step (iteration).
Using a constraint around the present robot location, the particle filter provides initial hypotheses with a uniform sampling of probable robot locations across the environment. For each unique particle, that is, the signal source, the Gaussian probability is determined. The particles are subsequently given weights that are proportional to their likelihood, and the weights \(w_i\)
are normalized as:
\begin{equation}
    w_i^*(q_i)=\frac{w_i(q_i)}{\sum_{i=0}^{N-1}w_i(q_i)}.
\end{equation}
This normalized weight determines the likelihood that a particle from this set of particles will be regenerated. The particle with the highest weight is the most recent best gauge of the mobile robot's location. This process is repeated until either one unique molecule remains or no modern tests are available. It is worth noting that the PF is iterated for each unused estimation tuple. Alg.~\ref{alg:main} provides an overview of the proposed method. All these computations are being performed by the moving vehicle, which is running a centralized service and receives the RSSI information from all connected nodes that sense wireless signals independently in a synchronized manner.

 \begin{algorithm}[t]
 \SetAlgoLined
  Pr = [] \% Initial List of Particles in the PF \;
  \While{end of trajectory stream}{
  sample \(x_t\) from transition model \(P(X_{t+1}|x_t)\)\;
   \(X_{t+1} = X_t\) \% motion model for robot\;
   W = [] \% weights\;
   \For{\(x_{t+1}\) in Pr}{
   add \(P(e_{t+1}|x_{t+1})\) to W\;
   }
   Rr = [] \% re-sampled particle filter\;
   \For{n=2 to N}{
   Choose P = Pr[n] of probability $w_i(x_i)$ (Eq.~\eqref{eqn:weights}) \;
   add P to Rr\;
   }
   Pr = Rr\;
   \(x_l^* = x_i \in w_i(x_i)\) is \(max(w_i)\)\;
  }
  \caption{PF Localization of mobile robot using DOA from WSNs}
  \label{alg:main}
 \end{algorithm}

\section{Experiments}
We implemented our approach and compared the performance with relevant recent methods from the literature on both extensive simulation studies and public real-world wireless signal datasets.
We have made 100 trials for each experiment and got the localization error as the distance between the predicted position and the actual position under normal circumstances. 

\paragraph*{Simulation Experiments}
We performed extensive simulations for verifying the performance of the proposed localization in terms of accuracy measured through the Root Mean Squared Error (RMSE) and efficiency measured through the Time Per Iteration (TPI).

\begin{figure*}
    \centering
\begin{subfigure}{.64\textwidth}
\centering
\hspace{-8mm}
\includegraphics[width=\linewidth]{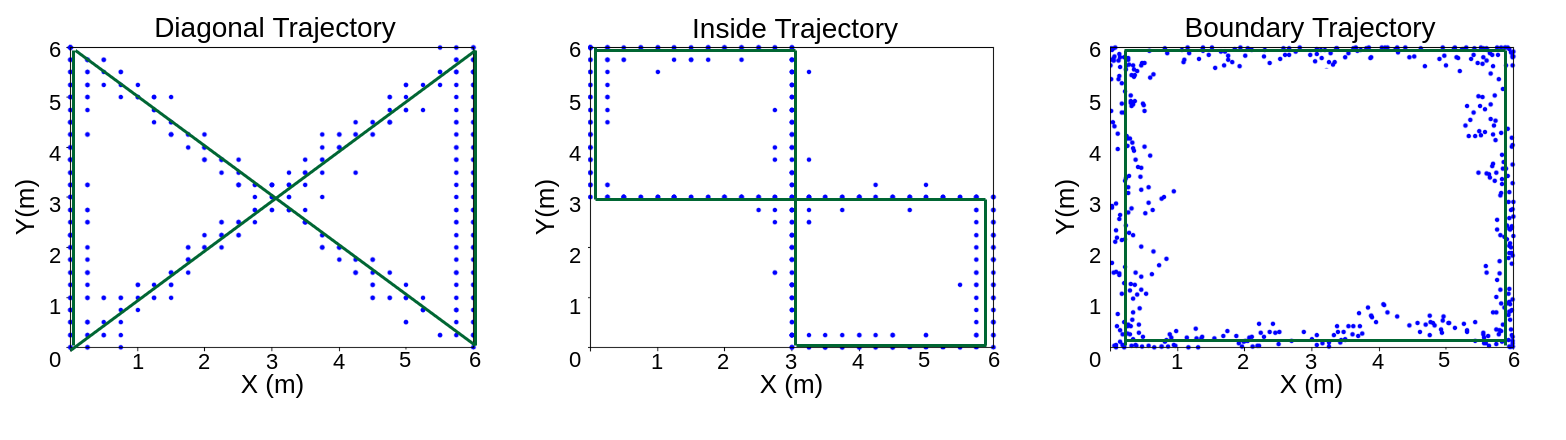}
\vspace{-4mm}
\caption{Test trajectories in simulations (Dataset 0).}
\label{simulation_trajectory}
\end{subfigure}
\begin{subfigure}{.17\textwidth}
\centering
\includegraphics[width=\linewidth]{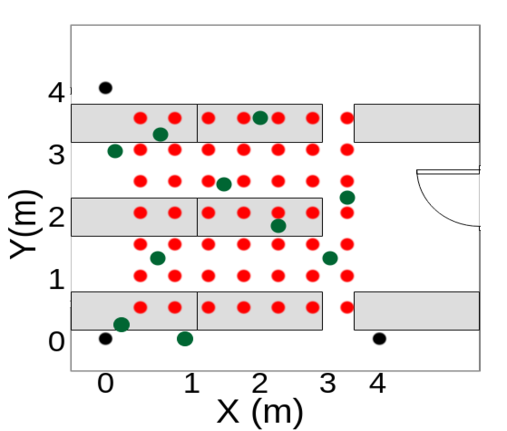}
\caption{Dataset 1 \cite{dataset1}.}
\label{dataset1}
\end{subfigure}
\begin{subfigure}{.17\textwidth}
\centering
\includegraphics[width=\linewidth]{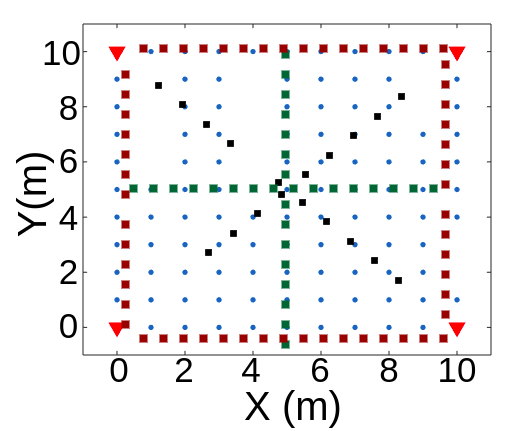}
\vspace{-5mm}
\caption{Dataset 2 \cite{dataset2}.}
\label{dataset2}
\end{subfigure}
\caption{The test samples in simulations and real-world datasets used in this study are shown here. The simulation plots show the actual trajectory (where the signal sample was taken) of the robot  in solid lines and the predicted locations in scattered dots. The dataset plots show the bounded environment with different points where signal sample and location information are available to test and compare our methods.}
    \label{fig:my_label}
    \vspace{-2mm}
\end{figure*}

\textbf{Dataset 0} We generated a simulated dataset in Python, for which we have used 4 WSNs distributed on the corners of the simulation workspace, with the initial position of the robot being their center (Fig.~\ref{fig:overview}). We simulated three different trajectories for robot motion: Boundary (left), Cross Coverage (center), Diagonal (right), and the scale of the workspace is 6m x 6m as shown in Fig.~\ref{simulation_trajectory}.
Through these paths, we cover all potential positions within the bounded region.
We measure the RSS value of each WSN node for all positions along the path of the robot. The estimated RSS based on log-normal radio signal fading model can be computed as:
\begin{equation}
    RSSI = A - 10\times n \times \log_{10}(d),
    \label{eq:rssi}
\end{equation}
where \(n\) denotes the path loss exponent, which varies between 2 (free space) and 6 (complex indoor environment), \(d\) denotes the distance from Robot \(R\) to the node \(N\), and \(A\) denotes received signal strength at a reference distance of one meter.
We used this dataset to perform experiments to validate the accuracy of localization techniques for different noise conditions on the measurements simulated through a zero-mean Gaussian noise varying from 1 to 4 dBm variance.

\paragraph*{Real-world Datasets}
\label{dataset}
We used two different publicly available real-world RSSI datasets on indoor localization: Dataset 1\footnote{\url{https://github.com/pspachos/RSSI-Dataset-for-Indoor-Localization-Fingerprinting}} provides RSSI values for three wireless technologies; BLE, Zigbee, and Wi-Fi in a 6m x 6m room ((Fig.~\ref{dataset1}); and Dataset 2\footnote{\url{https://ieee-dataport.org/documents/multi-channel-ble-rssi-measurements-indoor-localization}} provides RSSI values for the three regions of varying range in the bounded area: diagonal, boundary, and inside in a 10m x 10m room (Fig.~\ref{dataset2}). More details are available in Appendix~\ref{sec:appendix-dataset}.

\paragraph*{Compared methods}
To validate the results of our proposed approach, we implemented the four model-based solutions: 1) \textbf{trilateration} \cite{fundamentalTrilateration1996,yang2020trilateration}, 2) \textbf{weighted centroid localization} \cite{weightedCentroid2014survey}, 3) \textbf{differential RSSI-based localization}\cite{podevijn2018comparison}, 4) \textbf{Markov grid} approach which applies the same Gaussian probability over DOA approach as our PF but the Markov grid searches through each and every location on the workspace with a resolution of 0.5m (this is similar to the Expectation Maximization (EM) method in \cite{measRSS}).
More details on each of these methods are in Appendix~\ref{sec:appendix-methods}.

\begin{table*}[t]
\begin{center}
\caption{Results on both simulated and real-world datasets, showing accuracy (RMSE) and efficiency (TPI).}
\label{tab:results}
\vspace{-2mm}
\resizebox{\textwidth}{!}{
\begin{tabular}{|c|c|c|c|c|c|c|c|c|c|m{10em}|m{8em}|}
\hline
\multicolumn{2}{|c|}{}& \multicolumn{2}{|c|}{\textbf{Trilateration}}&\multicolumn{2}{|c|}{\textbf{WCL}}&\multicolumn{2}{|c|}{\textbf{D-RSSI}}&\multicolumn{2}{|c|}{\textbf{Markov}}&\multicolumn{2}{|c|}{\textbf{Proposed PF-DOA}} \\
\cline{3-12}
\multicolumn{2}{|c|}{\textbf{Experiment Basis}}
 & \textbf{\textit{RMSE (m)}} & \textbf{\textit{TPI (ms)}}
  & \textbf{\textit{RMSE (m)}} & \textbf{\textit{TPI (ms)}} 
   & \textbf{\textit{RMSE (m)}} & \textbf{\textit{TPI (ms)}} 
    & \textbf{\textit{RMSE (m)}} & \textbf{\textit{TPI (ms)}} 
     & \textbf{\textit{RMSE (m)}} & \textbf{\textit{TPI (ms)}} 
 \\
\hline
\multicolumn{12}{|l|}{Dataset 0 (Simulations)}\\
\hline
 \multicolumn{2}{|c|}{Diagonal} & 1.03$\pm$0.45 &71$\pm$20  &1.36$\pm$0.76 &61$\pm$18 &0.42$\pm$0.09 &\textbf{42$\pm$17} &\textbf{0.08$\pm$0.01} &105$\pm$29 &0.10$\pm$0.01 wo/ odom  0.08$\pm$0.01  w/ odom &49$\pm$13 wo/ odom 66$\pm$11 w/ odom
 \\
\cline{3-12}
\multicolumn{2}{|c|}{Inside} &1.41$\pm$0.57  & 68$\pm$18 &1.43$\pm$0.81 & 53$\pm$21 &0.53$\pm$0.12 &\textbf{52$\pm$13} &\textbf{0.09$\pm$0.015} &99$\pm$33 &0.12$\pm$0.08 wo/ odom  0.10$\pm$0.07 w/ odom & 53$\pm$14 wo/ odom  65$\pm$21 w/ odom
 \\
\cline{3-12}
 \multicolumn{2}{|c|}{Boundary} &1.23$\pm$0.68  &77$\pm$16  &1.62$\pm$0.96 &63$\pm$23 &0.72$\pm$0.18 &\textbf{48$\pm$16} &\textbf{0.21$\pm$0.09} & 112$\pm$38&0.36$\pm$0.08 wo/ odom  0.26$\pm$0.07 w/ odom & 51 $\pm$12 wo/ odom  62$\pm$18 w/ odom
  \\
\cline{3-12}
 \multicolumn{2}{|c|}{Average} &1.22$\pm$0.56  &72$\pm$18  &2.47$\pm$0.84 &59$\pm$20 &0.55$\pm$0.13 &\textbf{47$\pm$15} &\textbf{0.13$\pm$0.04} & 105$\pm$100&0.19$\pm$0.05 wo/ odom  0.15$\pm$0.05 w/ odom & 51 $\pm$13 wo/ odom  64$\pm$16 w/ odom
  \\
\hline
\multicolumn{12}{|l|}{Dataset 1}\\
\hline
 \multicolumn{2}{|c|}{Wi-Fi} &2.74$\pm$1.79  & \textbf{69$\pm$18} &3.83$\pm$1.85 &76$\pm$21 &2.21$\pm$1.24 &89$\pm$25 &1.52$\pm$0.54 &202$\pm$75 &\textbf{1.07$\pm$0.46} &95$\pm$30
 \\
\cline{3-12}
 \multicolumn{2}{|c|}{BLE} &2.33$\pm$1.62  &75$\pm$16  &3.76$\pm$0.81 & 83$\pm$17&1.90$\pm$0.91 &\textbf{69$\pm$10} &1.32$\pm$0.07 & 195$\pm$79&\textbf{1.18$\pm$0.24}&72$\pm$20
 \\
\cline{3-12}
  \multicolumn{2}{|c|}{Zigbee} &3.03$\pm$1.93  &78$\pm$14 & 4.21$\pm$0.21& \textbf{65$\pm$12} &2.42$\pm$1.64 & 79$\pm$15&1.89$\pm$0.98 &215$\pm$98 & \textbf{1.26$\pm$0.74}&82$\pm$25
  \\
\cline{3-12}
 \multicolumn{2}{|c|}{Average} &2.70$\pm$1.78  &74$\pm$16 & 3.94$\pm$0.95& \textbf{73$\pm$16} &2.18$\pm$1.26 & 79$\pm$17&1.58$\pm$0.53 &204$\pm$ 84 & \textbf{1.17$\pm$0.48}&83$\pm$25
  \\
\hline
\multicolumn{12}{|l|}{Dataset 2}\\
\hline
  & Diagonal &2.91$\pm$1.67 &82$\pm$21  &3.81$\pm$2.02 & 94$\pm$29&2.01$\pm$1.14 &\textbf{78$\pm$11} &1.42$\pm$0.09 & 270$\pm$113&\textbf{0.39$\pm$0.08} &120$\pm$30
 \\
\cline{3-12}
  Channel 0 & Inside & 3.05$\pm$1.89 &  91$\pm$18& 4.31$\pm$2.21& \textbf{79$\pm$22}& 2.72$\pm$1.08 &83$\pm$18 &1.01$\pm$0.34 & 310$\pm$102& \textbf{0.82$\pm$0.14}&135$\pm$35
 \\
\cline{3-12}
  & Boundary &3.93$\pm$2.01  &\textbf{86$\pm$20}  &4.72$\pm$2.47 &89$\pm$24 &2.92$\pm$1.95 &94$\pm$19 & 1.22$\pm$0.51&295$\pm$91 &\textbf{1.01$\pm$0.69} &115$\pm$39
 \\
\cline{3-12}
  & Average &3.31$\pm$1.85  &\textbf{86$\pm$20}  &4.28$\pm$2.23 &87$\pm$22 &2.55$\pm$1.39 &85$\pm$16 & 1.22$\pm$0.31 &291$\pm$102 &\textbf{0.74$\pm$0.30} &123$\pm$36
 \\
\cline{3-12}
  & Diagonal &3.03$\pm$2.41 & 91$\pm$22  & 4.13$\pm$2.92 & \textbf{84$\pm$31} & 2.91$\pm$1.37 & 98$\pm$11 & \textbf{1.72$\pm$0.61} & 310$\pm$129&1.89$\pm$0.81 & 112$\pm$35
 \\
\cline{3-12}
  Channel 0-39 & Inside & 4.13$\pm$1.95 &  \textbf{76$\pm$12}& 5.02$\pm$3.64& 82$\pm$29 & 2.71$\pm$1.14 & 87$\pm$24 &1.33$\pm$0.76 & 356$\pm$111& \textbf{1.26$\pm$0.92}& 129$\pm$37
 \\
\cline{3-12}
  & Boundary &4.02$\pm$2.47  & 79$\pm$22  & 5.62$\pm$3.24 & 83$\pm$23 & 3.12$\pm$ 2.05 & \textbf{74$\pm$19} & 1.92$\pm$0.84 & 374$\pm$105 &\textbf{1.85$\pm$0.76} &133$\pm$43
 \\
\cline{3-12}
  & Average &3.73$\pm$2.27  & \textbf{82$\pm$18}  & 4.93$\pm$3.26 & 83$\pm$27 & 2.92$\pm$ 1.52 & 86$\pm$18 & \textbf{1.66$\pm$0.73} & 346$\pm$115 &1.67$\pm$0.83 &122$\pm$38
 \\
\cline{3-12}
  & Diagonal &3.22$\pm$1.21 &\textbf{84$\pm$34}  &4.98$\pm$2.45 & 92$\pm$35&3.12$\pm$2.01 &89$\pm$23 &1.42$\pm$0.09 & 270$\pm$113&\textbf{1.99$\pm$0.41} &132$\pm$41
 \\
\cline{3-12}
  Channel 39 & Inside & 5.22$\pm$2.63 &  95$\pm$21& 6.41$\pm$2.16& 92$\pm$29& 4.12$\pm$3.21 &\textbf{84$\pm$22} &1.71$\pm$0.94 & 325$\pm$114& \textbf{1.64$\pm$1.04}&153$\pm$41
 \\
\cline{3-12}
  & Boundary &4.92$\pm$3.02  &91$\pm$29  &6.23$\pm$3.62 & \textbf{79$\pm$20} &4.21$\pm$2.03 & 96$\pm$29 & \textbf{2.02$\pm$1.21} & 372$\pm$112 &2.11$\pm$1.19 &145$\pm$49
 \\
\cline{3-12}
 & Average &4.45$\pm$2.28  &90$\pm$27  &5.87$\pm$2.74 & \textbf{87$\pm$28} &3.82$\pm$2.41 & 89$\pm$24 & \textbf{1.72$\pm$0.74} & 323$\pm$113 &1.91$\pm$0.88 &143$\pm$45
 \\
\hline
\end{tabular}
}
\end{center}
\vspace{-4mm}
\end{table*}

\section{Results and Discussion}
\label{sec:results}

\begin{figure*}
\centering
\begin{subfigure}{.36\textwidth}
 \includegraphics[width=\linewidth]{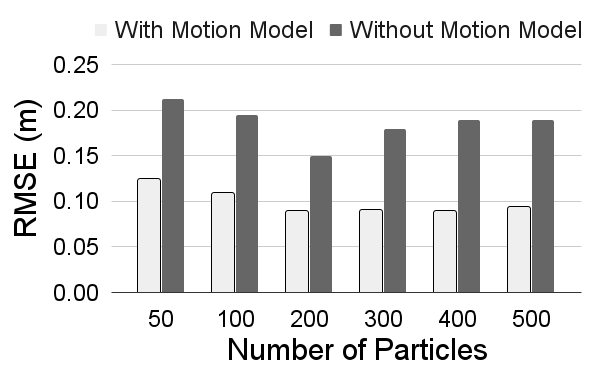}
 \caption{Number of Particles}
 \label{fig:particles}
\end{subfigure}
\begin{subfigure}{.36\textwidth}
 \includegraphics[width=\linewidth]{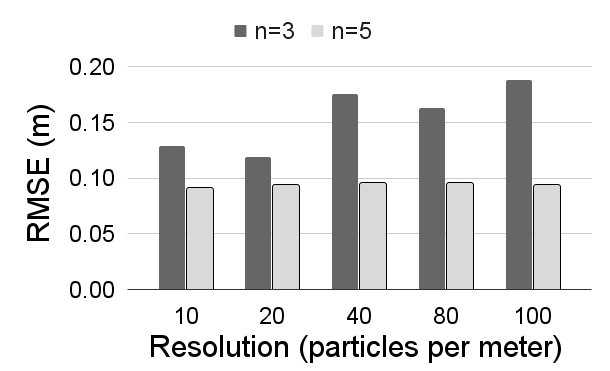}
 \caption{Grid Resolution}
 \label{fig:resolution}
\end{subfigure}
 \begin{subfigure}{.25\textwidth}
 \includegraphics[width=\linewidth]{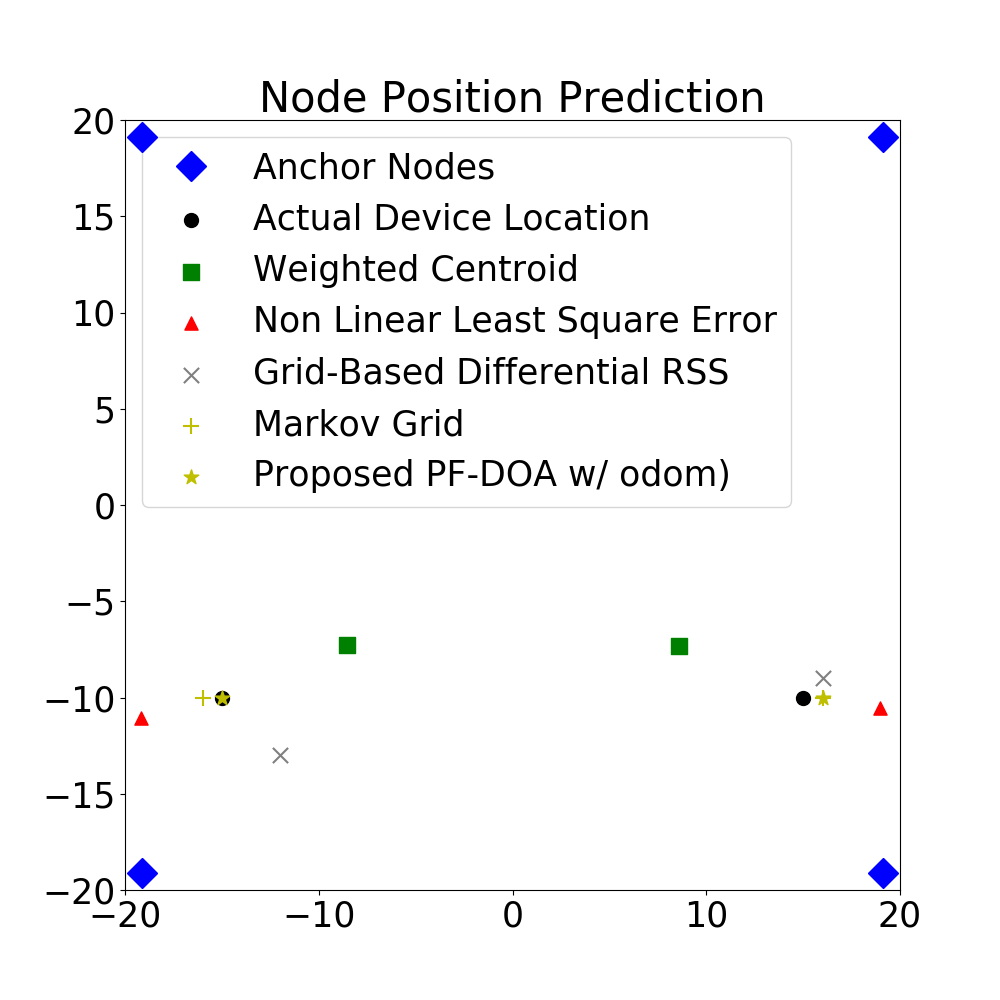}
 \caption{Scatter Plot}
 \label{fig:scatter}
\end{subfigure}
 \vspace{-2mm}
\caption{Factors affecting accuracy of position prediction in terms of RMSE. Left: Impact by the number of particles and inclusion of robot motion (odometry) model; Center: Impact by different propagation constants ($n$) and grid resolution; Right: Scattered position predictions in 40 x 40 m bounded region by different techniques.}
 \label{sim-results}
 \vspace{-3mm}
\end{figure*}

In the proposed approach, certain factors (number of particles, path-loss exponent, and resolution of particles) are involved that can impact the accuracy of localization. So, we find the optimal combination of such parameters for our method before comparing it with the methods from the literature.
Fig.~\ref{fig:particles} presents the results of the RMSE variations under different numbers of particles. 
It can be seen that by increasing the number of particles, the accuracy has improved as expected. However, it also increased the computational complexity, so we chose 200 as the optimal number of particles in the PF. 
By implementing the motion model along with the particle filter, convergence slightly improved the accuracy at a small expense in the computation time. This shows the utility of fusing the motion model (or IMU sensor) of the mobile robot with our PF-DOA localization.

The propagation constant ($n$ in Eq.~\eqref{eq:rssi}) plays a significant role in wireless signals since it provides an indication of the environment complexity in propagating wireless signals. Contrary to the belief that the higher the propagation constant, the lower the accuracy, our method performed better even under high $n$ values, as can be seen in Fig.~\ref{fig:resolution}. This is mainly because of the small boundary space of the experiment settings within which a larger propagation constant would produce larger variations in RSS, therefore high accuracy in localization. 
Similar to the impact on the number of particle filters, the increase in the particles' resolution also improves the accuracy at the cost of computation efficiency. We chose a path loss exponent of 3 in our further simulations to present reasonable complexity in the environment. Also, based on the data, we set the optimal PF resolution of 0.1 m for the PF method in our further analysis.

\begin{figure}[ht]
\centering
\vspace{-5mm}
\begin{center}
 \includegraphics[width=1.1\columnwidth]{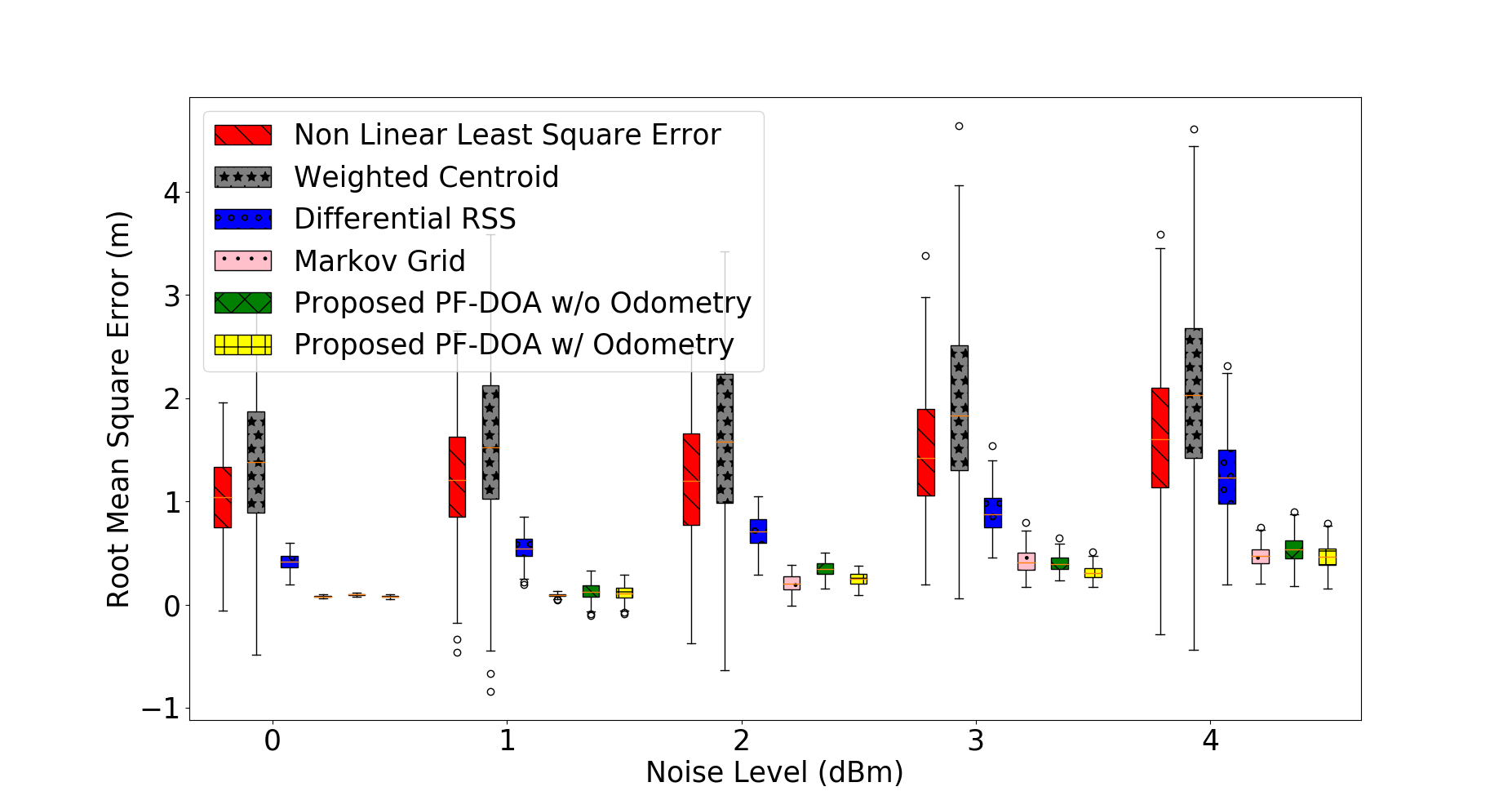}
\end{center}
\vspace{-4mm}
 \caption{Performance under different RSS noise levels.}
 \label{simulation_noise_performance_plot}
 \vspace{-5mm}
\end{figure}

\subsubsection{Simulation Results}
Table~\ref{tab:results} summarizes the results of all the experiments. In general, the proposed method performed significantly better than the state-of-the-art methods in terms of both accuracy and efficiency.

\paragraph*{Comparative picture} To further understand how our method localizes compared to other methods, we plotted the position predictions in a large simulated area of 40 m x 40 m bounded region. Fig.~\ref{fig:scatter} visualizes the localization accuracy for all methods in a few samples locations. We can observe that the proposed method predicted mobile device location very close to the actual location, even in a large area.

\paragraph*{Impact of signal noise}
We analyze the {performance} based on the accuracy in terms of RMSE for different simulated noise levels in the measured RSS. It can be seen in Fig.~\ref{simulation_noise_performance_plot} that among all techniques for different noise levels, the proposed approach has lower RMSE (high accuracy). The trilateration approach performed better than the Weighted Centroid method. However, both of them have 3x higher RMSE than the proposed and Markov method for most of the experiments. Markov and the proposed approach performed slightly better in terms of accuracy, but Markov approach is computationally complex as it calculates the Gaussian probability in all the areas of the grid with a full resolution instead of sparsely and randomly distributed particles in the PF.

\subsubsection{Real Data Results}
In general, our method outperformed other methods in both datasets. Further analysis for individual categories are detailed below.

\paragraph*{Wireless technology comparison}
Dataset 1 has RSSI observations captured from three different technologies (Wi-Fi, BLE, and Zigbee). The findings from these experiments Table.~\ref{tab:results} delineate the suitability of using Wi-Fi as a communication channel and the proposed approach for indoor localization as it has the least RMSE than other technologies with 1.07m RMSE in a 4m x 4m of bounded region. As expected, there is no significant difference in terms of computational complexity among different technologies because the methods take the same time to run the algorithm irrespective of where the signals are coming from. 
The proposed approach also has 0.4 m better accuracy than the KNN as presented in \cite{sadowski2020memoryless} for scenario 1 of the Dataset 1.

\paragraph*{Regional comparison}
Dataset 2 has RSSI  for three different regions in a bounded area of dimension 10 x 10 m. From Table.~\ref{tab:results} it can be evident that the proposed approach works well for the positions inside the AP/WSN perimeter compared to diagonal or boundary points. This is due to the fact that the DOA estimates can be ambiguous on the diagonal and boundary regions, resulting in higher localization error. However, in reality, the mobile device would be inside the infrastructure boundaries, and this is advantageous for our method.

\paragraph*{Channel comparison}
Dataset2 has RSSI observations for 40 different channels under three different regions in the bounded area of dimension 10 x 10 m.
We show the results for the channel with less noise (channel 0), the channel with high noise (channel 39) and report the combined RMSE for channels 0-39. It can be seen from Table.~\ref{tab:results} that the proposed approach consistently provided the best performance in most of the scenarios with reasonable computational efficiency. 

\subsubsection{Discussion}
The proposed PF-DOA approach is able to localize a mobile device using an existing WSN or AP infrastructure with up to 8 cm accuracy in simulations and meter-level of accuracy in real-world datasets. Obviously, the simulated dataset is created using a standard radio signal generator model with accurately known parameters and distance information. Hence, it has comparatively less uncertainty and noise in measurement than real-world dataset. 
Nevertheless, the trends of the comparison results between different methods hold for the real measurements as well.
Further, the PF-DOA approach performed better than the Markov approach in real-world datasets mainly because the resolution of the PF-DOA is 0.1 m (finer) compared to the 1 m (coarser) resolution of the Markov approach (limited by the sparse test data points available at this resolution). In comparison, both Markov and PF methods use 0.1 m as resolution in simulation experiments.

\textit{Limitations:} Similar to any wireless node collaboration-based approach \cite{wang2019assistant}, the proposed approach only works for more than three fixed nodes placed at geometrically-aligned positions of a regular polygon bounded region to obtain accurate DOA. While it is robust for most scenarios, it is dependent on the quality of the RSS and the obstacles or non-line of sight conditions, which need to be studied further.

\section{Conclusion}
We proposed a new model-based online localization method for mobile wireless devices in indoor environments. The approach employs a particle filter with a Bayesian resampling process using the Gaussian probability of wireless signal DOA, which is estimated through the geometric properties of the wireless sensor nodes. Considerable simulation experimentation has revealed promising results on the method's accuracy and efficiency. 
Our method outperformed by up to 5x improvement in accuracy compared to standard methods from the literature and showed remarkable efficiency in achieving accurate meter-level localization compared with the Markov localization approach, which can provide high accuracy but at the cost of computation time.

\bibliography{ehsan_bib,pfdoa_bib}
\bibliographystyle{IEEEtran}

\clearpage

\appendices

\section{Dataset Information}
\label{sec:appendix-dataset}
\textbf{Dataset 1} Provides RSSI values for three wireless technologies; BLE, Zigbee, and Wi-Fi. We have used the data for Scenario 1 (Fig.~\ref{dataset1}) as it relates to our approach in the way of geometric positioning of anchor nodes. In this scenario, a room of 6.0 x 5.5 m was used as the experiment testbed. 
All transmitting devices were removed from the surroundings to establish a clear testing medium where all devices could communicate without interference. 
The transmitters were spaced 4 meters apart in the shape of a triangle. 
The fingerprint and test points were obtained with a 0.5 m distance between the transmitters in the center. 
The database would be made up of 49 fingerprints as a result of this. Ten test points were chosen at random for testing.
We have arranged the fingerprinting dataset in such a way that it makes a trajectory in the region. 

\textbf{Dataset 2} Provides RSSI values for the three regions of varying range in the bounded area: diagonal, boundary, and cross-inside (Fig.~\ref{dataset2}). Four anchors took RSSI measurements while receiving messages from a single mobile node, delivering advertisement and extended advertisement messages in all BLE channels (both primary and secondary advertisement channels). Four anchors were placed in the corners of a 10 x 10 m office area (no large impediments).
We have compared the results for different communication channels under different regions in the bounded area.

\section{Compared Methods}
\label{sec:appendix-methods}
\textbf{Trilateration} \cite{fundamentalTrilateration1996}
Trilateration is a model-based technique that uses distances to determine the receiver's location numerically. To calculate with trilateration, we need three transmitting devices to obtain a 2-D position and four to find a 3-D position. The distances between the transmitter and the receivers, in addition to the right number of transmitting devices, are necessary. A frequent method for calculating the distance between devices is to use the RSSI of a signal. 
For 2-D space, with three anchor nodes $N_1,N_2,N_3$ and positions in space be $(a_1,b_1),(a_2,b_2),(a_3,b_3)$ respectively. We can find the unknown position $(x,y)$ of the receiver as:
 \begin{align*}&\begin{cases} \displaystyle (a_{1}-x)^{2}+(b_{1}-y)^{2}=d_{1}^{2} \\ \displaystyle (a_{2}-x)^{2}+(b_{2}-y)^{2}=d_{2}^{2} \\ \displaystyle (a_{3}-x)^{2}+(b_{3}-y)^{2}=d_{3}^{2} \end{cases} 
\end{align*}
To minimize the positing error, we need to minimize the following objective function using a non-linear least squares technique:
\begin{equation*} f(x,y)=\sum _{i=1}^{3}\left [{\sqrt {(x-a_{i})^{2}+(y-b_{i})^{2}}-d_{i}}\right]^{2}\end{equation*}

\textbf{Weighted Centroid}\cite{weightedCentroid2014survey}
The basic idea of a weighted centroid localization algorithm based on RSSI is that unknown nodes gather RSSI information from the beacon nodes around them.
Assuming there are n anchor nodes in the WSN, with coordinates $(x_1, y_1)$, $(x_2, y_2),. . .,(x_n, y_n)$, respectively, the location of the unknown node can be obtained by using the improved centroid algorithm estimating the coordinates of $n$ nodes as:
\begin{align*} &\begin{cases} x=\frac{{w_{1}}^{\ast}{x_{1}}^{+}{w_{2}}^{\ast}x_{2}{w_{3}}^{\ast}{x_{3} }^{+\ldots+}{w_{n}}^{\ast}x_{n}} {{w_{1}}^{+}{w_{1}}^{+}{w_{1}}^{+\ldots+}w_{1}}\\ y=\frac{{w_{1}}^{\ast}{y_{1}}^{+}{w_{2}}^{\ast}y_{2}{w_{3}}^{\ast}{y_{3} }^{+\ldots+}{w_{n}}^{\ast}y_{n}} {{w_{1}}^{+}{w_{1}}^{+}{w_{1}}^{+\ldots+}w_{1}}\\ \end{cases}\\ &w_{i}=\frac{RSSI_{i}}{RSSI_{1}+RSSI_{2}+RSSI_{3}+\ldots+RSSI_{n}}\\ &i\in(1,2,3,\ldots,n)\end{align*}

\textbf{Differential RSS}\cite{podevijn2018comparison}
The Differential RSS method works without knowing to transmit power beforehand. There are two phases in this technique; offline and online phases. During the offline phase, actual received RSS values are generated for each point of the grid using the representative (measured) RSS model. During the online phase, the measured DRSS values for each grid point are compared to the theoretical ones. The estimated location (X,Y) is determined as the grid point with the theoretical RSS values closest (least squares) to the ones measured:
\begin{equation*} (X,Y)=\min_{x,y}\sum_{i=1}^{N}(DRSS_{(x,y),i,T}-DRSS_{i,M})^{2} \tag{2} \end{equation*}
Here $DRSS_{(x,y),i,T}$ denotes the actual differential RSS value at position (x,y) from or at anchor i, $i=1 ... N$ with N is the number of anchors.$RSS_{i,M}$ is the measured RSS value from or at anchor i and is therefore required. $DRSS_0$ is obtained from a measurement at a reference point. DRSS can be calculated as:
\begin{equation*} DRSS_{i}=RSS_{i}-RSS_{1} \end{equation*}
Where $RSS_1$ denotes the strongest received signal strength. In the algorithm, $n$ is considered constant and known.

\textbf{Markov Grid} Markov's grid based localization uses an explicit, discrete representation for the probability of all positions in the state space. We represent the environment by a finite number of discrete state space (Grids). In the algorithm, at each iteration, the probability of each state of the entire space is updated. Use a fixed decomposition grid by discretizing each DOA: \((x, y, \theta)\). For each location \(x_i = [x,y,\theta]\) in the configuration space: determine probability \(P(x_i)\) of robot being in that state. Then, choose the state (position) with the highest probability. 
This approach is also similar to the Expectation Maximization (EM) method in \cite{measRSS}.

\end{document}